\documentclass{article}
\usepackage{spconf,amsmath,graphicx}
\usepackage{MnSymbol}
\usepackage{wasysym}
\usepackage{hyperref,url,latexsym,color,soul}

\title{From Speech-to-Speech Translation to Automatic Dubbing}
\name{Marcello Federico\qquad Robert Enyedi\qquad Roberto Barra-Chicote}
\secondlinename{Ritwik Giri\qquad Umut Isik\qquad Arvindh Krishnaswamy \qquad Hassan Sawaf$^\dag$\thanks{$\dag$ Contribution while the author was with Amazon.}}
\address{Amazon}
\begin{document}
\ninept
\maketitle
\begin{abstract}
We present enhancements to a  speech-to-speech translation pipeline in order to perform automatic dubbing. Our architecture features neural machine translation generating output of preferred length, prosodic alignment of the translation with the original speech segments, neural text-to-speech with fine tuning of the duration of each utterance, and, finally, audio rendering to enriches text-to-speech output with background noise and reverberation extracted from the original audio. We report on a subjective evaluation of automatic dubbing of excerpts of TED Talks from English into Italian, which measures the perceived naturalness  
of automatic dubbing and the relative importance of each proposed enhancement.
\end{abstract}
\section{Introduction}
\label{sec:intro}
Automatic dubbing can be regarded as an extension of the speech-to-speech translation (STST) task \cite{wahlster_verbmobil:_2013}, which is generally seen as the combination of three sub-tasks:  
(i) transcribing speech to text in a source language (ASR), (ii) translating text from a source to a target language (MT) and (iii) generating speech from text in a target language (TTS). 
Independently from the implementation approach \cite{weiss_sequence--sequence_2017,waibel_interactive_1996, vidal_finite-state_1997,metze_nespole!_2002,nakamura_atr_2006,casacuberta_recent_2008}, the main goal of STST is 
producing an output that reflects the linguistic content of the original sentence.
On the other hand, automatic dubbing aims to replace all speech contained in a video document 
with speech in a different language, so that the result sounds and looks as natural as the original. Hence, 
in addition to conveying the same content of the original utterance, dubbing should also match the original timbre, emotion, duration, prosody, 
background noise, and reverberation.  

While STST has been addressed for long time and by several research labs \cite{waibel_interactive_1996, vidal_finite-state_1997,metze_nespole!_2002,nakamura_atr_2006,wahlster_verbmobil:_2013},  relatively less and more sparse efforts have been devoted to automatic dubbing \cite{matousek_automatic_2010,matousek_improving_2012,furukawa_video_2016,oktem2019}, although the potential demand of such technology could be huge. In fact, multimedia content created and put online has been growing at exponential rate, in the last decade, while availability and cost of human skills for subtitling and dubbing still remains a barrier 
for its diffusion worldwide.\footnote{Actually, there is still a divide between countries/languages where either subtitling 
or dubbing are the preferred translation modes \cite{kilborn_`speak_1993,koolstra_pros_2002}. The reasons for this
are mainly economical and historical \cite{danan_dubbing_1991}.} Professional dubbing of a video file is a very labor 
intensive process that involves many steps: (i) extracting speech segments from the audio track and annotating these with speaker information; 
(ii) transcribing the speech segments, (iii) translating the transcript in the target language, (iv) adapting the translation for timing, 
(v) choosing the voice actors, (vi) performing the dubbing sessions, (vii) fine-aligning the dubbed speech segments, 
(viii) mixing the new voice tracks within the original soundtrack.

Automatic dubbing has been addressed both in  monolingual cross-lingual settings. 
In \cite{verhelst_werner_automatic_1997}, 
synchronization of two speech signals with the same content was tackled with time-alignment via dynamic time warping. 
In \cite{hanzlicek_towards_2008} automatic monolingual dubbing for TV users with special needs was generated from subtitles. 
However, due to the poor correlation between length and timing of the subtitles, TTS output frequently broke the timing boundaries.
To avoid unnatural time compression of TTS's voice when fitting timing constraints, \cite{matousek_automatic_2010} proposed phone-dependent time compression and 
text simplification to shorten the subtitles, while \cite{matousek_improving_2012} leveraged scene-change detection 
to relax the subtitle time boundaries.  Regarding cross-lingual dubbing, lip movements synchronization 
was tackled in \cite{furukawa_video_2016} by directly modifying the actor's mouth motion via shuffling of the 
actor's video frames. While the method does not use any prior linguistic or phonetic knowledge, 
it has been only demonstrated on very simple and controlled conditions. 
Finally,  mostly related to our contribution is \cite{oktem2019}, which discusses speech synchronization  
at the phrase level (prosodic alignment) for English-to-Spanish automatic dubbing. 


In this paper we present research work to enhance a STST pipeline in order to comply with the timing and rendering 
requirements posed by cross-lingual automatic dubbing of TED Talk videos. Similarly to \cite{matousek_automatic_2010}, we also shorten the TTS script by directly modifying the MT engine rather than via text simplification. As in \cite{oktem2019}, we synchronize phrases across languages, 
but follow a fluency-based rather than content-based criterion and replace generation and rescoring of hypotheses in \cite{oktem2019} with a more efficient dynamic programming solution. Moreover, we extend \cite{oktem2019} 
by enhancing neural MT and neural TTS to improve speech synchronization, and by performing audio rendering on the dubbed speech to make it sound more real inside the video.

In the following sections, we introduce the overall architecture (Section 2) and the proposed enhancements (Sections 3-6). Then, we present  results (Section 7) of experiments evaluating the naturalness of automatic dubbing  of TED Talk clips from English into Italian.  To our knowledge, this is the first work on automatic dubbing that integrates enhanced deep learning models for MT, TTS and audio rendering, and evaluates them on real-world videos. 





\begin{figure}[t]
    \centering
    \includegraphics[width=\columnwidth]{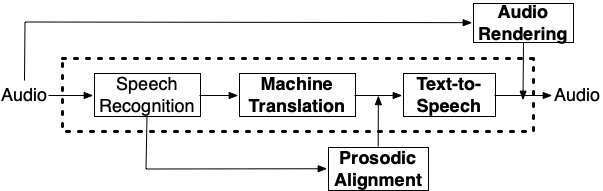}
    \caption{Speech-to-speech translation pipeline (dotted box) with enhancements to perform automatic dubbing (in bold).}
    \label{fig:NTTS}
\end{figure}

\section{Automatic Dubbing}
\label{sec:AD}

With some approximation, we consider here automatic dubbing of the audio track of a video as the task of STST, i.e. ASR + MT + TTS, with the additional requirement that the output must be temporally, prosodically and acoustically close to the original audio. We investigate an architecture (see Figure~1) that enhances the STST pipeline with 
(i) enhanced MT able to generate translations of variable lengths, (ii) a prosodic alignment module
that temporally aligns the MT output with the speech segments in the original audio, (iii) enhanced TTS to accurately control the duration of each produce utterance, and, finally, (iv) audio rendering that
adds to the TTS output background noise and reverberation  extracted from the original audio. In the following, we describe each component in detail, with the exception of ASR, for which we use \cite{DiGangi19} an of-the-shelf online service\footnote{Amazon Transcribe at https://aws.amazon.com/transcribe.}. ˜


\section{Machine Translation}
\label{sec:MT}
Our approach to control the length of MT output is inspired by \textit{target forcing} in multilingual neural MT \cite{johnson2016google, ha2016toward}. We partition the training sentence pairs into three groups (short, normal, long) according to the target/source string-length ratio. In practice, we select two thresholds $t_1$ and $t_2$, and partition training data according to the length-ratio intervals $[0,t_1)$, $[t_1,t_2)$ and $[t_2,\infty]$.  At training time a \textit{length token} is prepended to each source sentence according to its group, in order to let the neural MT model discriminate between the groups. At inference time, the length token is instead prepended to bias the model to generate a translation of the desired length type. We trained a Transformer model \cite{vaswani2017attention} with output length control on web crawled and proprietary data amounting to 150 million English-Italian sentence pairs (with no overlap with the test data).
The model has encoder and decoder with $6$ layers, layer size of $1024$, hidden size of $4096$ on feed forward layers, and $16$ heads in the multi-head attention.
For the reported experiments, we trained the models with thresholds $t_1=0.95$ and $t_2=1.05$ and generated at inference time translations of the shortest type, resulting in an average length ratio of $0.97$ on our test set. 
A detailed account of the approach, the followed training procedure and experimental results on the same task of this paper can be found in \cite{federico19}.  Finally, as baseline MT system we used an online service.~\footnote{Amazon Translate at https://aws.amazon.com/translate.}


\section{Prosodic Alignment}
\label{sec:PA}

Prosodic alignment\cite{oktem2019} is the problem of segmenting the target sentence 
to optimally match the distribution of words and pauses of the source sentence. Let ${\bf e}=e_1,e_2,\ldots,e_n$ be a source sentence of $n$ words which is segmented according to $k$ breakpoints $1 \le i_1 < i_2 < \ldots i_k=n$, shortly denoted with ${\bf i}$. Given a target sentence ${\bf f}=f_1,f_2,\ldots,f_m$ of $m$ words, the goal is to find within it $k$ corresponding breakpoints $1 \le j_1 < j_2 < \ldots j_k=m$ (shortly denoted with ${\bf j}$) that maximize the probability: 
\begin{equation}
\label{eq:1}
\max_{\bf{j}} \log \Pr({\bf j} \mid {\bf i},{\bf e},{\bf f})
\end{equation}
By assuming a Markovian dependency on ${\bf j}$, i.e.:
\begin{equation}
\Pr({\bf j} \mid {\bf i},{\bf e},{\bf f}) = \sum_{t=1}^k \log \Pr(j_t \mid j_{t-1}; t, {\bf i},{\bf e},{\bf f})
\end{equation}
and omitting from the notation the constant terms ${\bf i},{\bf e},{\bf f}$, we can derive the following recurrent quantity: 
\begin{equation}
Q(j,t) = \max_{j' < j} \log \Pr(j \mid j';t) + Q (j',t-1)    
\end{equation}
where $Q(j,t)$ denotes the log-probability of the optimal segmentation of ${\bf f}$ up to position $j$ with $t$ break points. It is easy to show that the solution of (\ref{eq:1}) corresponds to $Q(m,k)$ and that  
optimal segmentation can be efficiently computed via dynamic-programming. Let ${\tilde f}_t = f_{j_{t-1}+1},\ldots,f_{j_t}$ and  ${\tilde e}_t =e_{i_{t-1}+1},\ldots,e_{i_t}$ indicate the $t$-th segments of ${\bf f}$ and ${\bf e}$, respectively,  
we define the conditional probability of the $t$-th break point in ${\bf f}$ by: 
$$
\Pr(j_{t} \mid j_{t-1},t) \propto  \exp\left({1-\frac{|d(\tilde{e}_t)-d(\tilde{f}_t)|}{d(\tilde{e}_t)}}\right) \times \Pr(\mbox{br} \mid j_{t}, {\bf f})
$$
The first term computes the relative match in duration between the corresponding $t$-th segments\footnote{We approximate the duration $d(\cdot)$ of a segment with the sum of the lengths of its words. We plan to use better approximations in the future, e.g. the number of syllables \cite{oktem2019}.}, while the second term measure
the linguistic plausibility of a placing a break after the  ${j_t}$ in ${\bf f}$. For this, we simply compute the following 
ratio of language model perplexities computed over a text window centered on the break point, by assuming or not the presence of a pause (comma, semicolon or dash) in the middle: 
$$
\Pr(\mbox{br} \mid j, {\bf f}) = \frac{\Pr(f_{j},\mbox{br},f_{j+1})^{1/3}}{\Pr(f_{j},\mbox{br},f_{j+1})^{1/3}+\Pr(f_{j},f_{j+1})^{1/2}}
$$
In our implementation, we use a larger text window (last and first two words), we replace words with parts-of speech, and estimate the language model on a large English corpus.

\section{Text To Speech}
\label{sec:TTS}
Our neural TTS system consists of two modules: a Context Generation module, which generates a context sequence from the input text, and a Neural Vocoder module, which converts the context sequence into a speech waveform. The first one is an attention-based sequence-to-sequence network \cite{ntts01,ntts02} that predicts a Mel-spectrogram given an input text. A grapheme-to-phoneme module converts the sequence of words into a sequence of phonemes plus augmented features like punctuation marks and prosody related features derived from the text (e.g. lexical stress). For the Context Generation module, we trained speaker-dependent models on two Italian voices, male and female, with 10 and 37 hours of high quality recordings, respectively. We use the Universal Neural Vocoder introduced in \cite{tts03}, pre-trained with 2000 utterances per each of the 74 voices from a proprietary database.   

\noindent
To ensure close matching of the duration of Italian TTS output with timing information extracted from the original English audio, 
for each utterance we resize the generated Mel spectrogram using spline interpolation prior to running the Neural Vocoder. 
We empirically observed that this method produces speech of better quality than traditional time-stretching.

\section{Audio Rendering}
\subsection{Foreground-Background Separation}
The input audio can be seen as a mixture of foreground (speech) and background (everything else) and our goal is to 
extract the background and add it to the dubbed speech to make it sound more real and similar to the original.
For the foreground-background separation task, we adapted the popular U-Net \cite{ronneberger2015u} architecture, 
which is described in detail in \cite{jansson2017singing} for a music-vocal separation task. 
It consists of a series of down-sampling blocks, followed by one ’bottom’ convolutional layer, followed by a series
of up-sampling blocks with skip connections from the down-sampling to the up-sampling blocks. Because of the down-sampling blocks, the model can compute a number of high-level features on coarser time scales, which are concatenated with the local, high-resolution
features computed from the same-level up-sampling block. This concatenation results into multi-scale features for prediction.
The model operates on a time-frequency representation (spectrograms) of the audio mixture and it outputs two soft ratio masks corresponding to foreground and background, respectively, which are
multiplied element-wise with the mixed spectrogram, to obtain the final estimates of the two sources. Finally, the estimated spectrograms go through an inverse short-term Fourier transform block to produce raw time domain signals. The loss function used to train the model is the sum of the $L_1$ losses between the target and the masked input spectrograms, for
the foreground and the background \cite{jansson2017singing}, respectively. The model is trained with the Adam optimizer on mixed 
audio provided with foreground and background ground truths. Training data was created from 360 hours of clean speech from Librispeech (foreground) and 120 hours of recording taken from audioset \cite{gemmeke2017audio} (background), from which speech was filtered out using a Voice Activity Detector (VAD). Foreground and background are mixed for different signal-to-noise ratio (SNR), to generate the audio mixtures.


\subsection{Re-reverberation}
In this step, we estimate the environment reverberation from the original audio and apply it to the dubbed audio. Unfortunately, 
estimating the room impulse response (RIR) from a reverberated signal requires solving an ill-posed blind deconvolution problem. 
Hence, instead of estimating the RIR, we do a blind estimation of the reverberation time (RT), which is commonly used to assess 
the amount of room reverberation or its effects. The RT is defined as the time interval in which the energy of a steady-state 
sound field decays 60 dB below its initial level after switching off the excitation source. In this work we use a Maximum Likelihood 
Estimation (MLE) based RT estimate (see details of the method in \cite{lollmann2010improved}). Estimated RT is then used to 
generate a synthetic RIR using a publicly available RIR generator \cite{habets2006room}. This synthetic RIR is finally applied to 
the dubbed audio.

\section{Experimental Evaluation}
\label{sec:EE}
We evaluated our automatic dubbing architecture (Figure 1), by running perceptual evaluations in which users are asked to grade the naturalness of video clips dubbed with three configurations (see Table~\ref{tbl:systems}): (A) speech-to-speech translation baseline, (B) the baseline with enhanced MT and prosodic alignment, (C) the former system enhanced with audio rendering.\footnote{Notice that after preliminary experiments, we decided to not evaluate the configuration {\em A with Prosodic Alignment}, given its very poor quality, as also reported in \cite{oktem2019}.}
\begin{table}[h]
\label{tbl:systems}
\begin{center}
\caption{Evaluated dubbing conditions.}
\begin{tabular}{c|l}
System & Condition \\
 \hline
 R & Original recording (reference) \\ 
 \hline
 A      & Speech-to-speech translation (baseline) \\
 \hline
 B      & A with Enhanced MT and Prosodic Alignment \\
  \hline
 C      & B with Audio Rendering \\
 \hline
\end{tabular}
\end{center}
\end{table}
\noindent
Our evaluation focuses on two questions:
\begin{itemize}
    \item What is the overall naturalness of automatic dubbing?
    \item How does each introduced enhancement contribute to the naturalness of automatic dubbing? 
\end{itemize}

\begin{figure}[t]
    \centering
    \includegraphics[width=0.95\columnwidth]{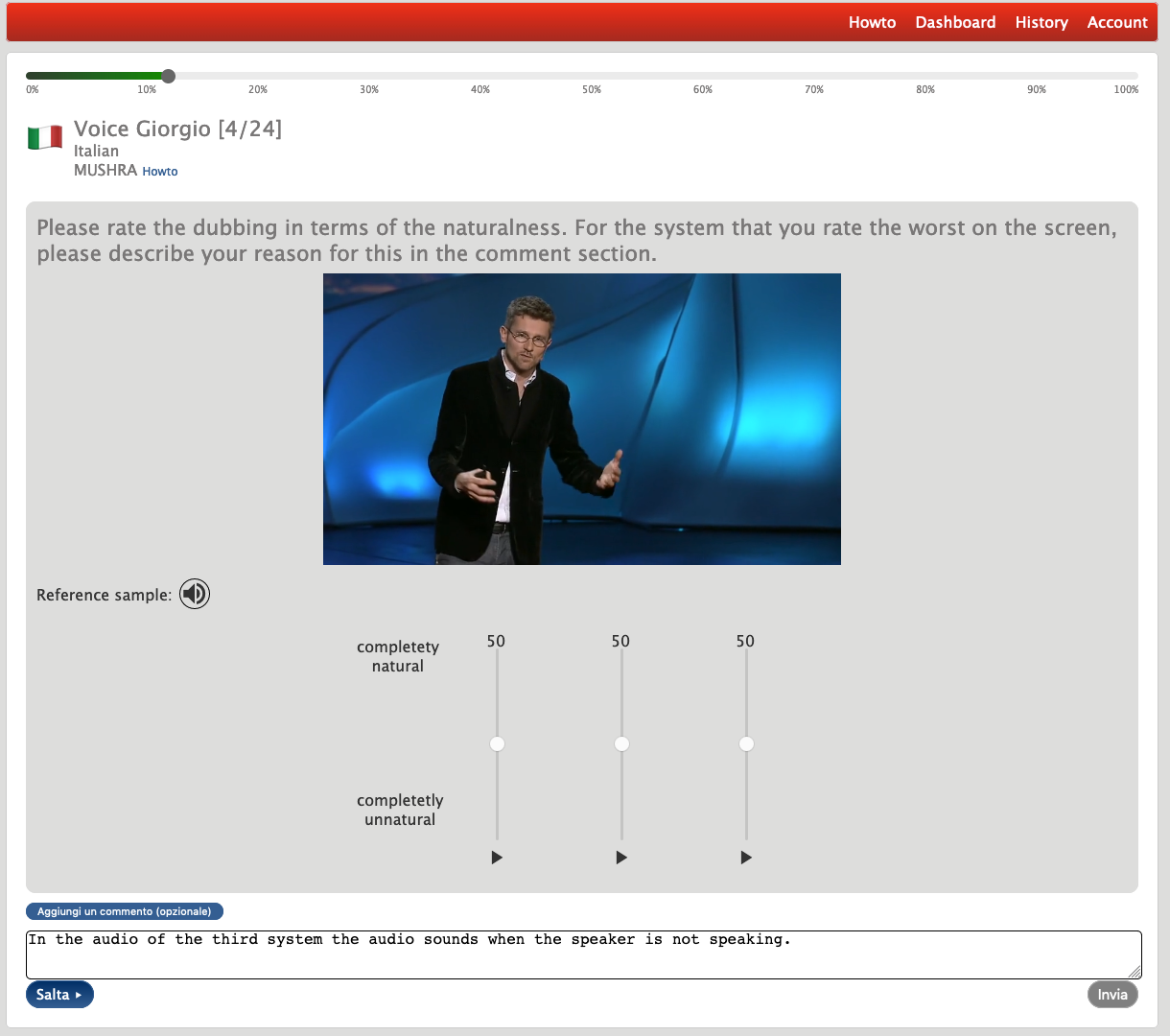}
    \caption{MUSHRA perceptual evaluation interface}
    \label{fig:MUSHRA platform}
\end{figure}

\noindent
We adopt the MUSHRA (MUlti Stimulus test with Hidden Reference and Anchor)  methodology \cite{mushra}, 
originally designed to evaluate audio codecs and later also TTS. We asked listeners to 
evaluate the naturalness of each versions of a video clip  on a 0-100 scale. Figure \ref{fig:MUSHRA platform} shows the user interface.  
In absence of a human dubbed version of each clip, we decided to use, for calibration 
purposes, the clip in the original language as hidden reference. The clip versions to evaluate are not labeled and randomly ordered. The observer has to play each version at least once before moving forward and can leave a comment about the worse version.   

In order to limit randomness introduced by ASR and TTS across the clips and by MT across
versions of the same clip, we decided to run the experiments using manual speech transcripts,
one TTS voice per gender, and MT output by the baseline (A) and enhanced MT system (B-C) of quality judged
at least acceptable by an expert.\footnote{We use the scale: 1 - Not acceptable: not fluent or not correct;  2 - Acceptable: almost fluent and almost correct; 3 - Good: fluent and correct.}  With these criteria in mind, we selected 24 video clips from 6 TED Talks 
(3 female and 3 male speakers, 5 clips per talk) from the official test set of the
MUST-C corpus \cite{di_gangi_must-c:_2019} with the following criteria: duration of around 10-15 seconds, 
only one speaker talking, at least two sentences, speaker face mostly visible. 

\noindent
We involved in the experiment both Italian and non Italian listeners. 
We recommended all participants to disregard the content and only focus on the naturalness 
of the output. Our goal is to measure both language independent and language dependent naturalness, i.e.  to verify how speech in the video resembles human speech with respect to acoustics and synchronization, and how intelligible it is to native listeners. 





\subsection{Results}
\label{sec:EV}
We collected a total of 657 ratings by 14 volunteers, 5 Italian and 9 non-Italian listeners, spread over the 24 clips and three testing conditions. 
We conducted a statistical analysis of the data with linear mixed-effects models using the {\tt lme4} package for R \cite{bates_fitting_2015}. We analyzed the naturalness score (response variable) against the following two-level fixed effects: dubbing system A vs. B, system A vs. C, and  system B vs. C. We run separate analysis for Italian and non-Italian listeners.  In our mixed models, listeners and video clips are random effects, as they represent a tiny sample of the respective true populations\cite{bates_fitting_2015}. We keep models maximal, i.e. with  intercepts and slopes for each random effect, end remove terms required to avoid singularities 
\cite{bates_parsimonious_2015}.  
Each model is fitted by maximum likelihood and significance of intercepts and slopes are computed via t-test. 

Table ~\ref{tab:mixed} summarized our results. In the first comparison, baseline (A) versus the system with enhanced MT and prosody alignment (B), we see that both non-Italian and Italian listeners perceive a similar naturalness of system A (46.81 vs. 47.22). When movid to system B,  non-Italian listeners perceive a small improvement (+1.14), although not statistically significant, while Italian speaker perceive a statistically significant degradation (-10.93).  In the comparison between B and C (i.e. B enhanced with audio 
rendering), we see that non-Italian listeners observe a significant increase in naturalness (+10.34), statistically significant, while Italian listeners perceive a smaller and not statistical significant improvement (+1.05). The final comparison between A 
and C gives almost consistent results with the previous two evaluations: non-Italian listeners perceive better quality in condition C
(+11.01) while Italian listeners perceive lower quality (-9.60). Both measured variations are however not statistically significant due to the higher standard errors of the slope estimates $\Delta$C. Notice in fact that each mixed-effects model is trained on distinct data sets and with different random effect variables. A closer look at the random effects parameters indeed shows  that for the B vs. C comparison, the standard deviation estimate of the listener intercept is 3.70, while for the A vs. C one it is 11.02. In other words, much higher variability across user scores is observed in the A vs. C case rather than in the B vs. C case. 
A much smaller increase is instead observed across the video-clip random intercepts, i.e. from 11.80 to 12.66. 
The comments left by the Italian listeners tell that the main problem of system B is the unnaturalness of the speaking rate, i.e. is is either too slow, too fast, or too uneven. 

\begin{figure}[t]
    \centering
    \includegraphics[width=1\columnwidth]{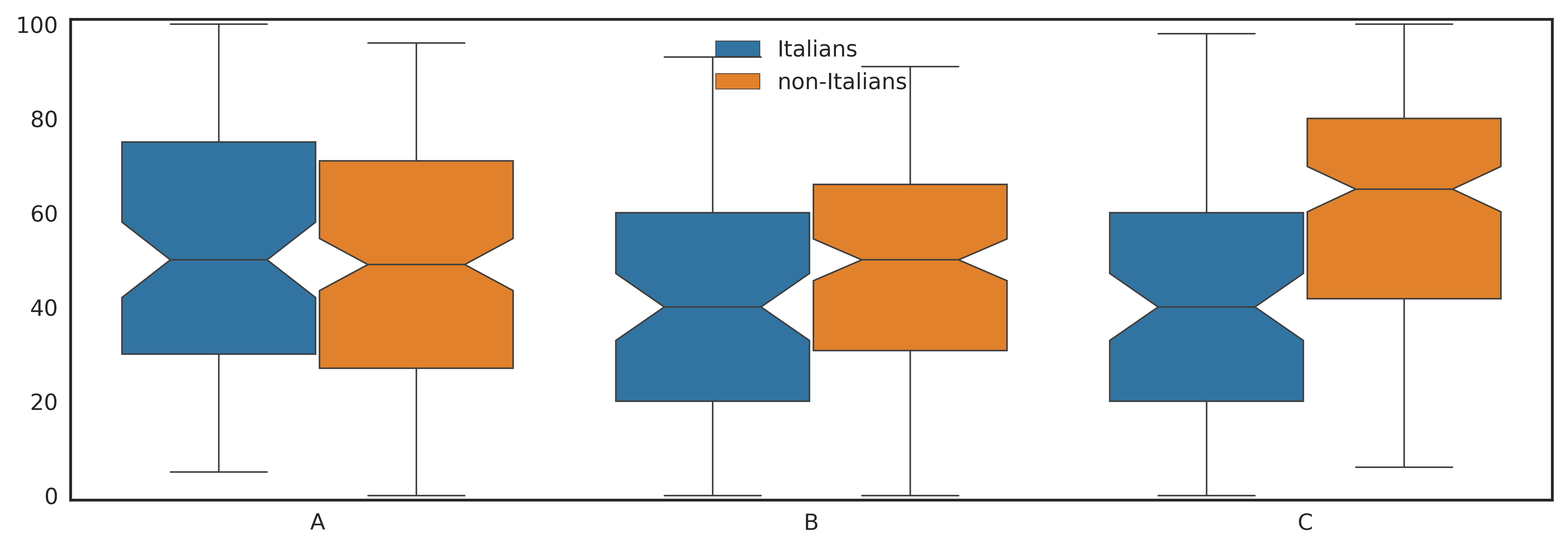}
    \includegraphics[width=1\columnwidth]{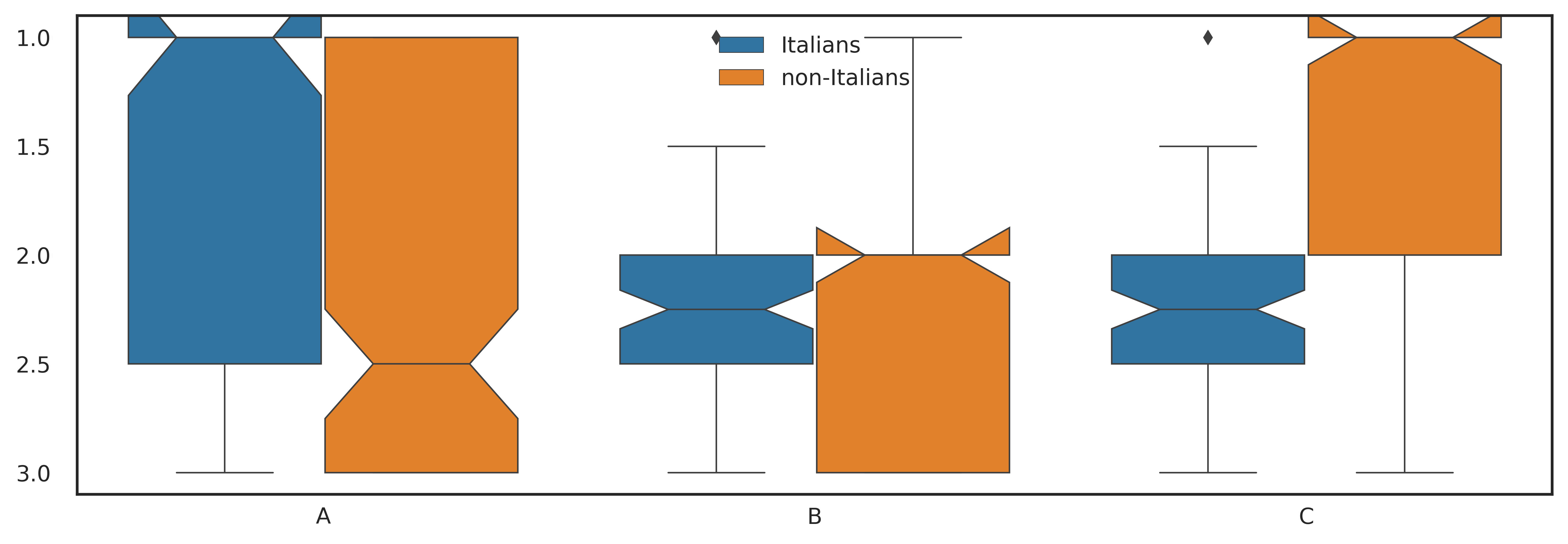}
    \caption{Boxplots with the MUSHRA scores (top) and Rank Order (bottom) per system and mother language (Italian vs Non-Italian).}
    \label{fig:MUSHRA evaluation results}
\end{figure}

The distributions of the MUSHRA scores presented at the top of Figure~\ref{fig:MUSHRA evaluation results} confirm our analysis. What is more relevant, the distribution of the rank order (bottom) strengths our previous analysis. Italian listeners tend to rank system A the best system (median $1.0$) and vary their preference between systems B and C (both with median $2.0$). In contrast, non-Italian rank system A as the worse system (median $2.5$), system B as the second (median $2.0$), and statistically significantly prefer system C as the best system (median $1.0$).


\begin{table}[]
    \centering
    \begin{tabular}{l|r|r|r|r|}{}
      & \multicolumn{2}{c|}{Non Italian} & \multicolumn{2}{c}{Italian}\\
      
      Fixed effects & Estim & SE & Estim. & SE \\
      \hline
      $A$ intercept & 46.81$^\bullet$  &4.03 & 47.22$^\bullet$ & 6.81 \\
      $\Delta B$ slope & +1.14$^{\ }$ & 4.02 & -10.93$^*$ & 4.70  \\
       \hline
       \hline
     $B$ intercept & 47.74$^\bullet$ & 3.21 &  35.19$^\bullet$ & 7.22 \\
      $\Delta C$ slope & +10.34$^+$ & 3.53 &  +1.05$^{\ }$ & 2.30  \\
        \hline
      \hline
      $A$  intercept & 46.92$^\bullet$ & 4.95 &  45.29$^\bullet$ & 7.42  \\
      $\Delta C$ slope & +11.01$^{\ }$ & 6.51 &  -9.60$^{\ }$ & 4.89  \\
     \hline
    \end{tabular}
    \caption{Summary of the analysis of the evaluation with mixed-effects models. From top down: A vs. B, B vs. C, A vs. C. For each fixed effect, we report the estimate and standard error. Symbols $^\bullet$, $^*$, $^+$ indicate significance levels of 0.001, 0.01, and 0.05, respectively.}
    \label{tab:mixed}
\end{table}


Hence, while our preliminary evaluation found that shorter MT output can potentially enable better synchronization, the combination of MT and prosodic alignment appears to be still problematic and prone to generate unnatural speech. 

The incorporation of audio rendering (system $C$) significantly improves the experience of the non-Italian listeners (66 in median) respect to systems $B$ and $C$. This points out the relevance of including para-linguist aspects (i.e. applause's, audience laughs in jokes,etc.) and acoustic conditions (i.e. reverberation, ambient noise, etc.). For the target (Italian) listeners this improvement appears instead masked by the disfluencies introduced by the prosodic alignment step. If we try to directly measure the relative gains given by audio rendering, we see that Italian listeners score system B better than system A 27\% of the times and system C better than A 31\% of the times, which is a 15\% relative gain. On the contrary non-Italian speakers score B better than A 52\% of the times, and C better than A  66\% of the times, which is a 27\% relative gain.


\section{Conclusions}
We have perceptually evaluated the impact on the naturalness of automatic speech dubbing when we enhance a baseline speech-to-speech translation system with the possibility to control the length of the translation output, align target words with the speech-pause segmentation of the source, and enrich speech output with ambient noise and reverberation extracted from the original audio. We tested our system with both Italian and non-Italian listeners in order to evaluate both language independent and language dependent naturalness of dubbed videos. Results show that while we succeeded at achieving synchronization at the phrasal level, our prosodic alignment step negatively impacts on the fluency and prosody of the generated language. The impact of these disfluencies on native listeners seems to partially mask the effect of the audio rendering with background noise and reverberation, which instead results in a major increase of naturalness for non-Italian listeners. Future work will definitely devoted to improving the prosodic alignment 
component, by computing better segmentation and introducing more flexible lip-synchronization. 

\section{Acknowledgements}
The authors would like to thank the Amazon Polly, Translate and  Transcribe research teams; Adam Michalski, Alessandra Brusadin, Mattia Di Gangi and Surafel Melaku for   contributions to the project, and all colleagues at Amazon AWS who helped with the evaluation.



\ninept
\bibliographystyle{ieeetr}
\bibliography{biblio}
\end{document}